\documentclass[sigconf]{acmart}
\AtBeginDocument{%
  }

\copyrightyear{2026}
\acmYear{2026}
\setcopyright{cc}
\setcctype{by}

\acmConference[MM '26]
{Proceedings of the 34th ACM International Conference on Multimedia}
{November 10--14, 2026}
{Rio de Janeiro, Brazil}

\acmBooktitle{Proceedings of the 34th ACM International Conference on Multimedia
(MM '26), November 10--14, 2026, Rio de Janeiro, Brazil.}

\acmISBN{979-8-4007-2213-4/2026/11}
\acmDOI{10.1145/3767308.3836118}

\usepackage{multirow}
\graphicspath{{pics/}{../pics/}}
\usepackage{booktabs}
\usepackage[table]{xcolor}
\usepackage{xcolor}

\begin{document}

\title{NanoMorph-3D: An End-to-End Physics-Driven Unrolling Framework for Nanomaterial Reconstruction}

\author{Beiyuan Zhang}
\email{byzh@bit.edu.cn}
\affiliation{%
  \institution{Beijing Institute of Technology}
  \city{Beijing}
  \country{China}}

\author{Hesong Li}
\email{lihesong2@bit.edu.cn}
\affiliation{%
  \institution{Beijing Institute of Technology}
  \city{Beijing}
  \country{China}}

\author{Ziqi Wu}
\email{wuziqi@bit.edu.cn}
\affiliation{%
  \institution{Beijing Institute of Technology}
  \city{Beijing}
  \country{China}}

\author{Ruiwen Shao}
\email{rwshao@bit.edu.cn}
\affiliation{%
  \institution{Beijing Institute of Technology}
  \city{Beijing}
  \country{China}}

\author{Ying Fu}
\authornote{Corresponding author.}
\email{fuying@bit.edu.cn}
\affiliation{%
  \institution{Beijing Institute of Technology}
  \city{Beijing}
  \country{China}}

\renewcommand{\shortauthors}{Zhang et al.}


\begin{abstract}
Precise 3D characterization of nanomaterials is essential for unlocking structure-property relationships. However, standard electron tomography is fundamentally limited by the missing wedge problem. Consequently, conventional algorithms suffer from severe geometric distortions, a challenge further complicated by pervasive noise interference. Current learning-based methods either rely on physics-blind post-processing or employ end-to-end architectures constrained by local receptive fields, failing to capture complex 3D topologies. We propose NanoMorph-3D, a unified end-to-end framework grounded in a comprehensive Nanomorphological Taxonomy. Powered by a large-scale synthetic dataset explicitly modeling non-linear electron attenuation, we design a Physics-Driven Unrolled Network mapping proximal gradient descent into a learnable architecture. To capture complex internal topologies, we formulate a hierarchical attention mechanism with Physics-Normalization for long-range 3D dependencies and scale invariance. Crucially, our Dual-Domain strategy leverages Sinusoidal Attention to explicitly model physical projection trajectories, enforcing strict sinogram consistency to mitigate missing wedge artifacts. Finally, an unsupervised dual-stream mechanism bridges the simulation-to-reality gap. Experiments demonstrate NanoMorph-3D reconstructs diverse topologies with superior fidelity and speed.

\end{abstract}




\maketitle

\section{Introduction}
\label{sec:intro}

\begin{figure}[!tb] 
  \centering
  \includegraphics[width=0.95\columnwidth]{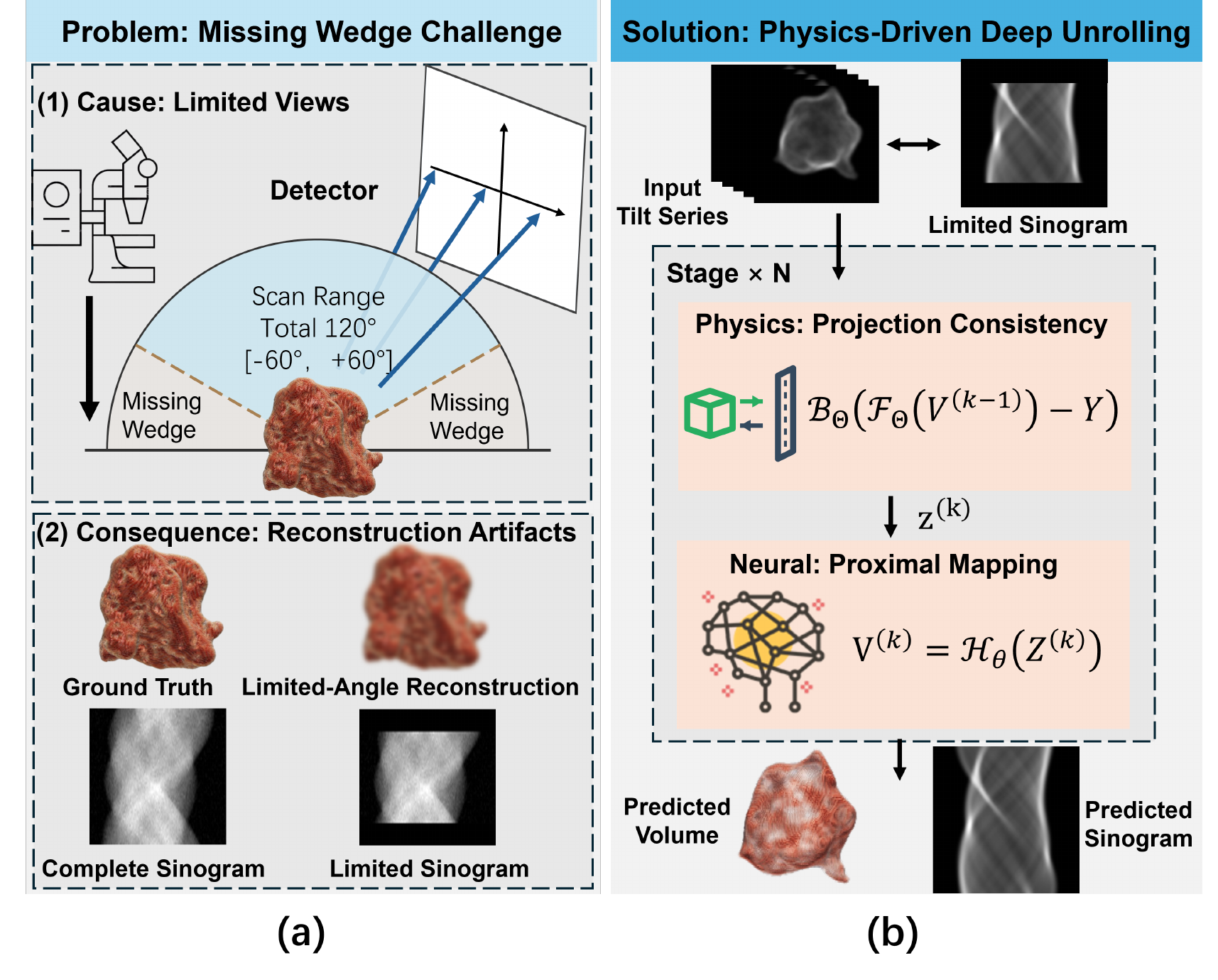} 
  \caption{\textbf{Overview of NanoMorph-3D.} 
  (a) \textbf{Missing Wedge Challenge:} Limited-angle acquisition leads to incomplete sinograms and anisotropic elongation artifacts in 3D reconstructions. 
  (b) \textbf{NanoMorph-3D:} A physics-driven deep unrolling architecture that recovers high-fidelity 3D morphologies and complete sinograms directly from input tilt series by coupling physical models with deep learning.}
  \Description{Overview diagram of the proposed NanoMorph-3D architecture.}
  \label{fig:teaser}
\end{figure}

Precise determination of three-dimensional mesoscale structures is fundamental to materials science because spatial morphology governs functional properties~\cite{tahir2020nanotechnology, ghasemi2024three, Nicolaides2001} such as catalytic activity and electronic mobility. Electron tomography~\cite{weyland2007electron, scott2012electron, Goris2011, Schwartz2022} is the premier technique for visualizing these structures. By acquiring a series of two-dimensional projections at different tilt angles, this method enables the mathematical reconstruction of three-dimensional volumes and provides critical insights into the internal architecture of complex nanomaterials.

High-fidelity reconstruction remains challenging due to the missing wedge problem \cite{gemmi20193d, ercius2015electron} and structural diversity. Unlike medical CT scans that acquire data over a full $360^{\circ}$ rotation~\cite{goldman2007principles}, electron tomography is inherently constrained by the mechanical structure of the microscope's sample stage. Moreover, at high tilt angles, the carbon support film and out-of-field materials occlude the specimen.

Consequently, data acquisition is typically restricted to a limited angular range, leaving a wedge-shaped region of Fourier space completely unsampled. As illustrated in Fig.~\ref{fig:teaser}(a), this incomplete sampling causes missing wedge artifacts that manifest as elongation and blurring along the electron beam direction, distorting fine topological features. Furthermore, nanomaterials exhibit a morphological spectrum ranging from dense nanoparticles to complex porous frameworks, making generalized reconstruction exceedingly difficult.

Existing solutions struggle to address these challenges concurrently. Conventional iterative algorithms~\cite{andersen1984simultaneous, trampert1990simultaneous, pryor2017genfire, pham2023accurate, Wei2023} rely on generic mathematical regularizers that over-smooth high-frequency details, often collapsing delicate internal topologies. Conversely, deep learning methods generally fall into two flawed paradigms. First, black-box post-processing modules~\cite{lee2021single} decouple denoising from the physical measurement model, leading to poor domain generalization and structural hallucinations. 

Second, end-to-end unrolled networks~\cite{gong2019mapem, adler2018learned, chen2024unsupervised} enforce physical constraints but do not explicitly model the projection geometry, leaving their spatial-domain prior unable to capture long-range 3D continuity.

To bridge this gap, we present NanoMorph-3D as a unified end-to-end framework. Central to our data foundation is a comprehensive Nanomorphological Taxonomy~\cite{darwish2024advancements, Sannino2021, Jahanian2024, Yang2023} that categorizes materials by intersecting macroscopic dimensions with internal textures, including dense, hollow, porous, and layered structures. Guided by this taxonomy, we construct a large-scale synthetic dataset. Crucially, our dataset explicitly models real-world non-linear electron attenuation via the Beer-Lambert law~\cite{swinehart1962beer} alongside realistic instrumental degradations. This rigorous physical testbed establishes a comprehensive dataset to evaluate generalizable topological recovery.

To this end, we present NanoMorph-3D. Our main contributions are as follows:
\begin{itemize}
    \item We propose NanoMorph-3D, an end-to-end framework that explicitly unrolls Proximal Gradient Descent (PGD) into a learnable physics-guided architecture. By formulating mechanisms to establish long-range 3D spatial dependencies and introducing Physics-Normalization, the model achieves robust, scale-invariant topological recovery with significantly accelerated inference.
    
    \item We introduce a Dual-Domain strategy empowered by Sinusoidal Attention to explicitly model projection trajectories. This strictly enforces sinogram consistency to eliminate missing wedge artifacts, while an unsupervised dual-stream mechanism enables seamless simulation-to-reality transfer on experimental observations.
    
    \item We construct a large-scale, physics-based 3D nanomaterial dataset driven by a systematic Nanomorphological Taxonomy. By rigorously modeling non-linear electron attenuation and instrumental degradations, we provide a comprehensive dataset to promote generalizable reconstruction research within the community.
\end{itemize}

\section{Related Work}
\label{sec:related}

\noindent\textbf{Tomographic Reconstruction Methods.} Conventional tomographic reconstruction relies heavily on analytical methods like Weighted Back-Projection~\cite{radermacher2006weighted} and iterative algorithms such as SIRT~\cite{trampert1990simultaneous} and Total Variation minimization~\cite{hu2014improved}. The accuracy and computational cost of iterative inversion also depend strongly on the fidelity of the physical forward model~\cite{li2019forwardModelsUSCT}. However, under limited-angle conditions, the missing wedge problem introduces anisotropic resolution and causes severe geometric distortions. Furthermore, generic handcrafted priors tend to penalize high-frequency variations, leading to over-smoothing that obliterates delicate internal topologies in complex porous nanomaterials. To address these limitations, data-driven deep learning approaches have been widely adopted. In related computational imaging settings, self-attention has been used to capture long-range dependencies and reconstruct images directly from under-sampled measurement sequences~\cite{tian2023transformerSPI}. Image-domain post-processing networks~\cite{liu2022isotropic, lee2021single, cciccek20163d} achieve high empirical precision but typically operate as physics-blind refinement modules. By decoupling the denoising process from raw physical measurements, they limit interpretability, struggle with domain generalization, and frequently hallucinate structures within unseen topologies. Concurrently, emerging computer vision techniques, such as implicit neural representations~\cite{mildenhall2021nerf, gao2022nerf} and explicit 3D Gaussian Splatting~\cite{kerbl20233d}, offer powerful 3D reconstruction capabilities. Yet, standard vision architectures are inherently designed for opaque surfaces. While recently adapted for tomographic transmission~\cite{ruckert2022neat, yu2025x2, qu2025gem}, their reliance on extensive per-scene optimization prohibits high-throughput screening. Furthermore, lacking explicit geometric constraints for the missing wedge, they struggle to disentangle internal features and frequently hallucinate within unsampled regions.

\begin{figure*}[t]
    \centering
    \includegraphics[width=\textwidth]{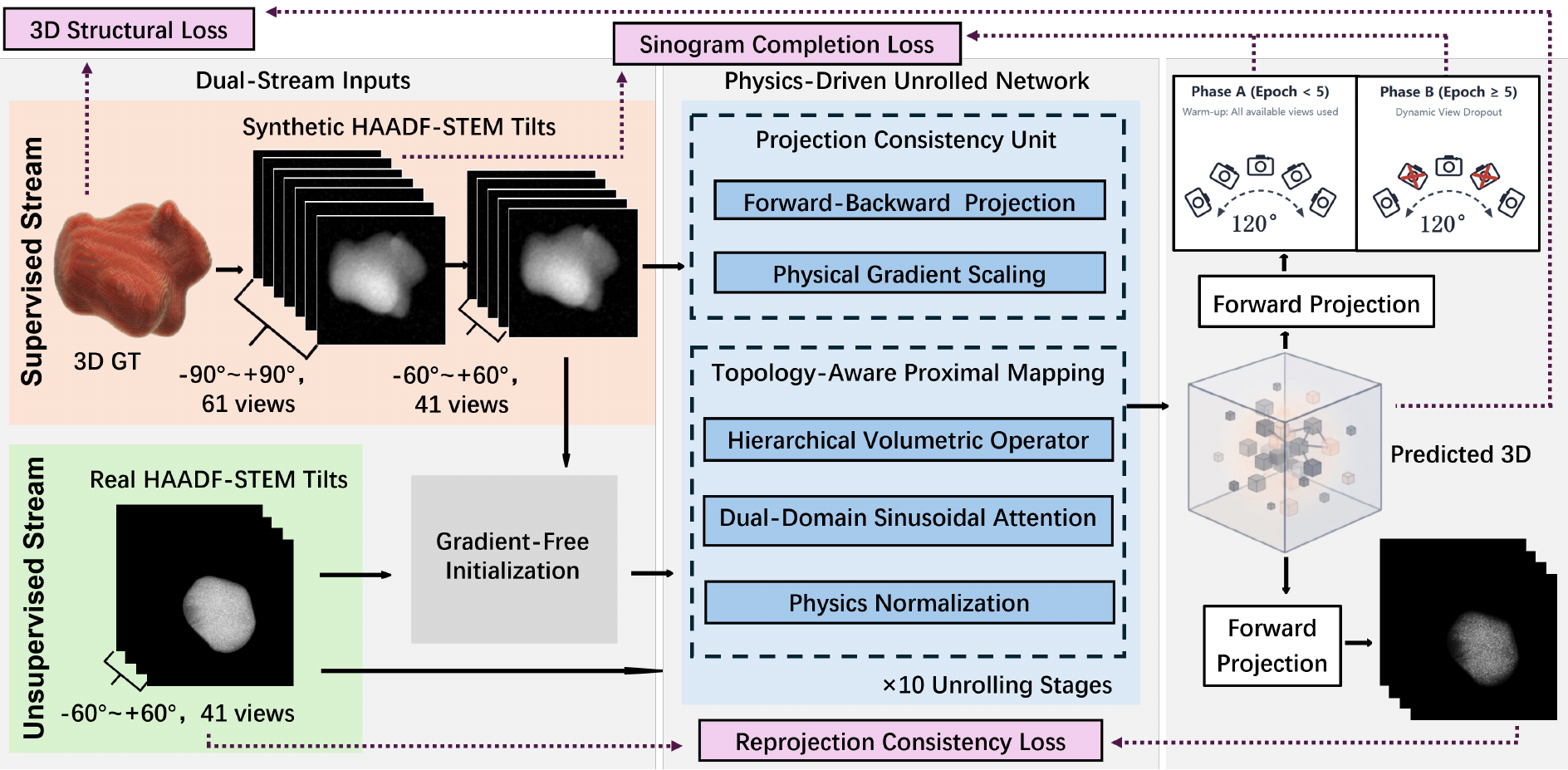}
    \caption{\textbf{Architecture of NanoMorph-3D.} Synthetic and real HAADF-STEM tilt series are processed by a dual-stream pipeline of ten unrolled stages, each alternating a Projection Consistency Unit with a Topology-Aware Proximal Mapping built from a Hierarchical Volumetric Operator, Dual-Domain Sinusoidal Attention, and Physics-Normalization. Training uses 3D structural and sinogram completion losses on synthetic data and a reprojection consistency loss on real data, with Dynamic View Dropout applied in Phase B.}
    \Description{Architecture diagram of the NanoMorph-3D dual-stream unrolled network.}
    \label{fig:framework}
\end{figure*}

\noindent\textbf{Deep Unrolling Networks.} Deep unrolling networks bridge the gap between iterative optimization and deep learning by unfolding algorithms, such as proximal gradient descent, into learnable layers. This paradigm maintains data consistency with the physical measurement model while leveraging the expressive power of neural networks. While promising, existing unrolling frameworks~\cite{liu2019deep, adler2018learned} predominantly rely on 2D or shallow 3D Convolutional Neural Networks. Constrained by localized receptive fields, their learned priors remain generic spatial-domain refinements that lack explicit geometric awareness of the anisotropic missing wedge and fail to model physical projection trajectories, rendering them incapable of establishing the long-range 3D structural continuity essential for complex nanomaterials. NanoMorph-3D advances this paradigm by formulating a hierarchical attention mechanism to explicitly model global context and structural connectivity. 
Rather than relying solely on spatial-domain refinement, we introduce a Dual-Domain Sinusoidal Attention mechanism to explicitly model physical projection trajectories. By tightly coupling deep proximal mapping with a targeted physics-driven trajectory model, our end-to-end framework offers a mathematically grounded and generalizable approach to the missing wedge problem.

\section{Method}
\label{sec:method}
\subsection{Overview}
Our objective is to reconstruct 3D nanomaterial volumes from noisy, limited-angle HAADF-STEM projections. To overcome signal degradation and missing wedge hallucinations, we propose NanoMorph-3D, a unified dual-domain framework shown in Fig.~\ref{fig:framework}. Sec.~\ref{subsec:problem_formulation} formulates electron tomography as an ill-posed inverse problem, and Sec.~\ref{subsec:pgd_unrolling} unrolls Proximal Gradient Descent into a physics-driven architecture. To resolve missing wedge ambiguity, Sec.~\ref{subsec:dual_domain} embeds Geometry-Aware Attention to enforce physical consistency across spatial and projection domains. To support this framework, Sec.~\ref{subsec:dataset_simulation} establishes a physics-grounded simulation dataset based on a Nanomorphological Taxonomy. Finally, Sec.~\ref{subsec:sim_to_real} introduces a dual-stream strategy enabling simulation-to-reality transfer on unlabeled experimental data.

\subsection{Problem Formulation}
\label{subsec:problem_formulation}
In HAADF-STEM imaging, the projection process is governed by the incoherent scattering of electrons. Unlike Atomic Electron Tomography which traces discrete atoms at scales below 10 nm, mesoscale tomography around 100 nm faces atomic overlapping along the projection path. Consequently, rather than following a linear mass-thickness assumption, the measured intensity $I$ at the detector undergoes non-linear exponential attenuation dictated by the Beer-Lambert law. By normalizing the intensity against the incident beam $I_0$ and the material attenuation coefficient $\mu$, the discrete forward physical model is formulated in the linear logarithmic domain as
\begin{equation}
    Y = -\frac{1}{\mu s} \ln\left(1-\frac{I}{I_0}\right) = \mathcal{A}_{\Theta} V + \eta
\end{equation}
where $s$ denotes the physical voxel size, $V \in \mathbb{R}^{N^3}$ represents the continuous three-dimensional mass density field, and $Y \in \mathbb{R}^{M \times N^2}$ is the stack of two-dimensional linearized projections acquired at $M$ distinct tilt angles $\Theta$. $\mathcal{A}_{\Theta}$ is the discrete Radon transform operator modeling the tomographic projection, and $\eta$ encapsulates physical degradations including Poisson shot noise. 

Due to the mechanical constraints of the microscope sample holder, the tilt angles $\Theta$ are confined to a limited range, typically $\pm 60^\circ$. This geometric restriction renders the system matrix $\mathcal{A}_{\Theta}$ rank-deficient, yielding the missing wedge problem. Mathematically, this creates a null space in the inversion process, leading to Z-axis elongation artifacts and topological ambiguity. Traditional iterative algorithms attempt to reconstruct $V$ by minimizing a regularized objective
\begin{equation}
    \hat{V} = \arg\min_V \frac{1}{2} \|Y - \mathcal{A}_{\Theta}V\|_2^2 + \lambda \Phi(V)
\end{equation}
where $\Phi(V)$ is a prior such as Total Variation. However, generic sparsity priors penalize high-frequency spatial variations. This leads to over-smoothing, which obliterates the internal topologies and pore-channel connectivity of mesoscale nanomaterials.

\subsection{Physics-Driven Unrolled Network}
\label{subsec:pgd_unrolling}
To transcend the limitations of handcrafted priors, we unroll the proximal gradient descent algorithm into a learnable architecture. To alleviate the memory footprint of volumetric gradient computation, the network initializes the density field through a gradient-free block before engaging the deep mapping. For the $k$-th fully unrolled stage, the reconstruction alternates between a physics-driven gradient step and a data-driven proximal mapping step.

\noindent\textbf{Projection Consistency Unit.} 
To enforce physical alignment with the raw measurements, the Projection Consistency Unit (PCU) performs a differentiable gradient update within each unrolled stage. Specifically, it constrains the intermediate volume $Z^{(k)}$ by back-projecting the residual error from the projection domain as
\begin{equation}
    Z^{(k)} = \text{ReLU}\left[V^{(k-1)} - \frac{\alpha^{(k)}}{M \cdot N} \mathcal{B}_{\Theta} \left[ \mathcal{F}_{\Theta}\left[V^{(k-1)}\right] - Y \right] \right]
\end{equation}
where $\mathcal{F}_{\Theta}$ represents the physical forward projection operator governing beam attenuation and $\mathcal{B}_{\Theta}$ denotes the corresponding geometric tomographic back-projection operator. The learnable step size $\alpha^{(k)}$ is normalized by the number of acquisition views $M$ and the spatial resolution $N$. This scale normalization ensures that the error gradients are distributed uniformly across the volumetric grid, preventing numerical instability during mixed-precision training.

\noindent\textbf{Topology-Aware Proximal Mapping} 
Instead of relying on a closed-form proximal operator, we parameterize the proximal mapping $\mathcal{H}_{\theta^{(k)}}$ using a deep neural network as
\begin{equation}
    V^{(k)} = \mathcal{H}_{\theta^{(k)}}\left[Z^{(k)}\right]
\end{equation}
Specifically, we implement this mapping through a hierarchical volumetric operator integrated with sinusoidal positional encodings. Unlike convolutional networks constrained by localized receptive fields, this operator-driven architecture captures long-range spatial dependencies to maintain global structural coherence. To prevent grid artifacts and ensure topological integrity, we employ a 3D Shifted Window mechanism. By alternating between 0-shift and 2-shift voxel displacements across successive layers, the network effectively bridges physical grid boundaries. This cross-window interaction stitches local grain boundaries and vacuum regions without the need for computationally expensive 3D masking, thereby preserving the macroscopic structural connectivity and complex pore-channel topologies of the nanomaterials.

\noindent\textbf{Physics-Normalization} Electron microscope measurements exhibit scale variations due to different physical magnifications and material mass densities. To address this, we introduce a Physics-Normalization layer within the transformer blocks. Operating as a density-aware root-mean-square normalization, it modulates the channel-wise feature distributions to align with absolute physical mass density scales. This module guarantees scale-invariant feature extraction and prevents the network from overfitting to projection intensities.

\subsection{Geometry-Aware Dual-Domain Learning}
\label{subsec:dual_domain}
Existing deep learning models hallucinate structures within missing wedge regions due to the absence of geometric constraints. To mitigate these topological hallucinations, we propose a dual-domain strategy that injects the geometry of the projection domain into the spatial volume.

\begin{figure}[t]
    \centering
    \includegraphics[width=0.9\linewidth]{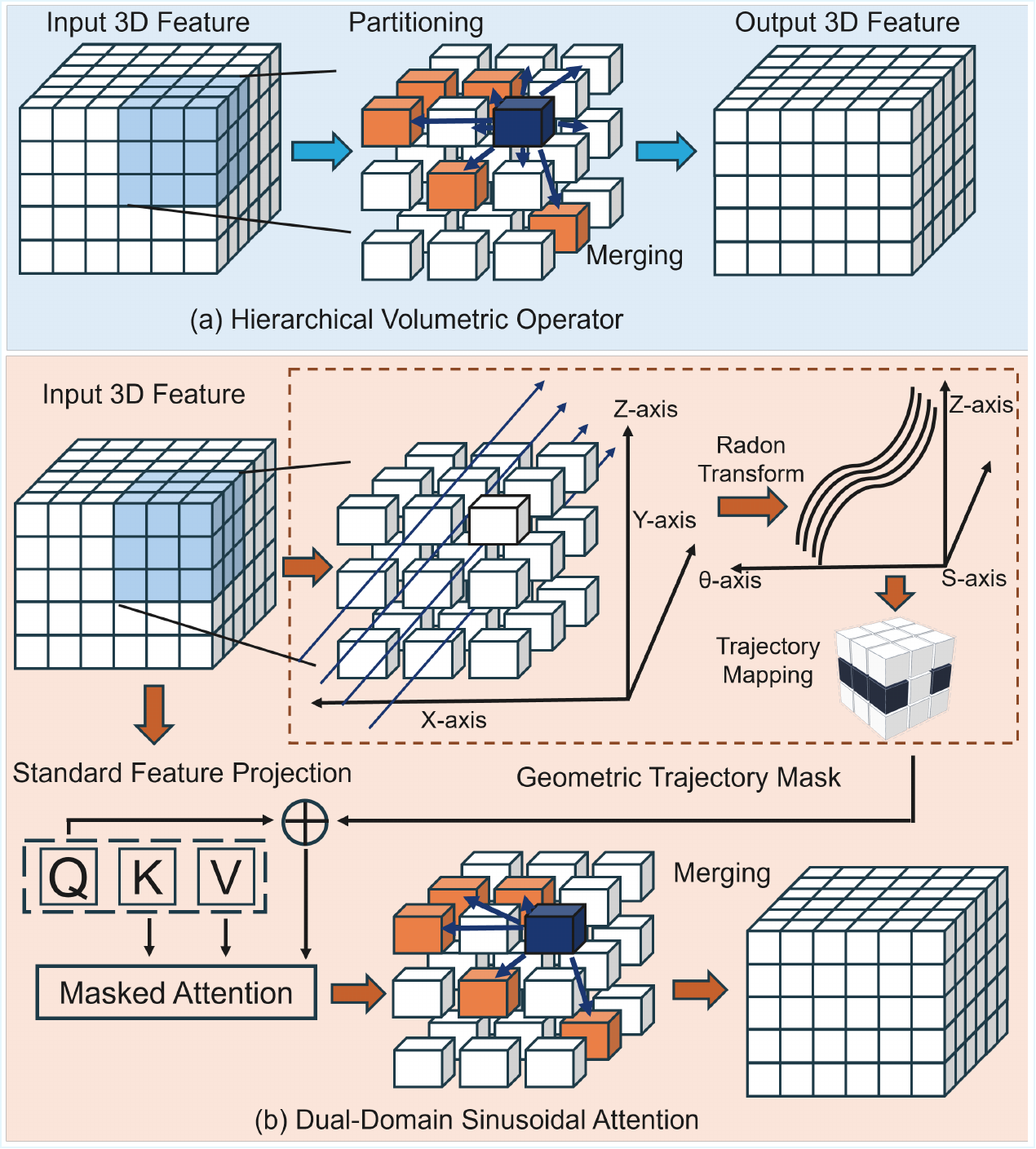}
    \caption{\textbf{Dual-Domain Sinusoidal Attention.} We bridge spatial and projection domains to construct a 3D geometric prior $\mathcal{M}_{\text{geo}}$ based on sinusoidal projection trajectories. Embedding this bias into the hierarchical operator guides feature exchange along physically valid trajectories to mitigate missing wedge artifacts and reduce topological hallucinations.}
    \Description{Diagram of the Dual-Domain Sinusoidal Attention.}
    \label{fig:attention}
\end{figure}

\noindent \textbf{Dual-Domain Sinusoidal Attention.} Standard volumetric operators process all voxels indiscriminately and ignore the anisotropic sampling density caused by the limited tilt range. As illustrated in Fig.~\ref{fig:attention}, to inject geometric awareness, we construct a pairwise trajectory-compatibility bias $\mathcal{M}_{\text{geo}}$. Under the Radon transform, a fixed 3D point traces a sinusoidal locus $u(\theta)=x\cos\theta+z\sin\theta$ across the tilt angles $\Theta$; $\mathcal{M}_{\text{geo}}$ encodes this sinogram geometry so that tokens exchange features along physically valid projection trajectories rather than by occupancy or spatial proximity alone. We inject it into the attention logits of the hierarchical operator as

\begin{equation}
\text{Output} = \text{Softmax}\left[ \frac{\mathbf{Q}\mathbf{K}^T}{\sqrt{d}} + \lambda\, g_\phi(\mathcal{M}_{\text{geo}}) \right]\mathbf{V}.
\end{equation}

Here $\lambda$ is a learnable scalar and $g_\phi$ a lightweight mapping. In this way, tokens exchange features along physically valid projection trajectories, allowing well-sampled regions to guide the completion of undersampled ones.

To overcome the spatial isolation in 3D windowed operations, we introduce a cyclic shifting mechanism. By alternating 3D window shifts across successive layers, the network stitches local topologies across grid boundaries to preserve the macroscopic integrity of complex mesoscale pore channels.

\noindent\textbf{Curriculum View Sampling Strategy.} To train the network to synthesize missing spatial frequencies without destabilizing early convergence, we introduce a curriculum-based Dynamic View Dropout strategy. In the initial warm-up phase (prior to the 5th epoch), the network receives the full limited-angle input to establish preliminary structural priors. For the supervised stream, this phase is strictly guided by a 3D Structural Loss ($\mathcal{L}_{\text{struct}}$) against the volumetric ground truth.

During the subsequent dynamic challenge phase, we randomly occlude a subset of projections within the available angular range. This mechanism effectively creates a self-supervised task. For synthetic data, the network must re-project the reconstructed volume $\hat{V}$ back to the target angles $\Theta_{\text{missing}}$, encompassing both the inherent static physical wedge and the dynamically dropped views, to compute a Sinogram Completion Loss ($\mathcal{L}_{\text{sino}}$) against the complete ground-truth sinogram:
\begin{equation}
    \mathcal{L}_{\text{sino}} = \|\mathcal{A}_{\Theta_{\text{missing}}}[\hat{V}] - Y_{\Theta_{\text{missing}}}\|_1
\end{equation}

\begin{figure*}[t]
    \centering
    \includegraphics[width=\textwidth]{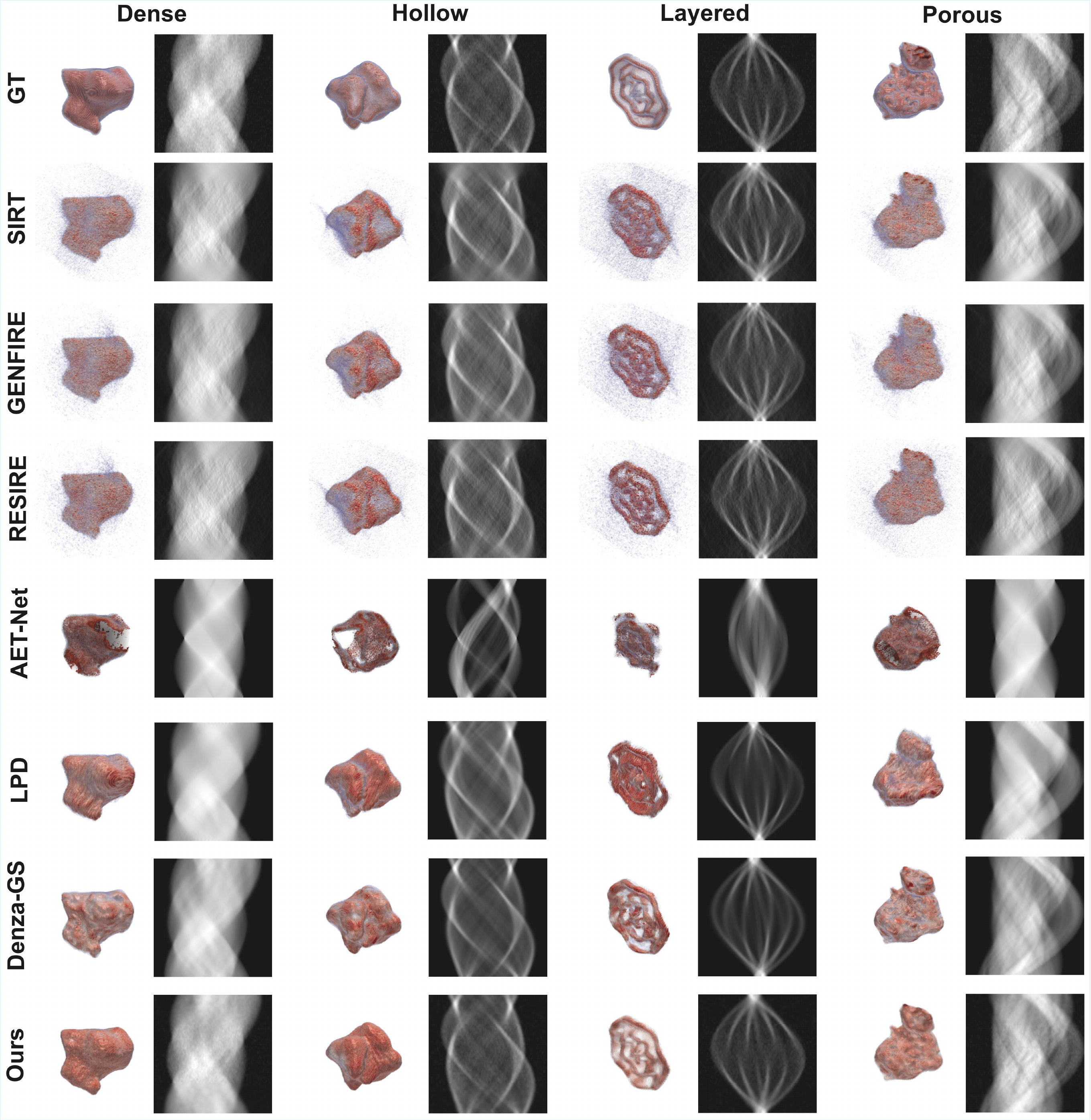}
    \vspace{-2mm}
    \caption{\textbf{Qualitative evaluation on the synthetic dataset.} Reconstructions of four morphological archetypes are shown with their sinogram slices. The iterative baselines SIRT, GENFIRE and RESIRE leave scattered noise artifacts, while the learned baselines AET-Net, LPD and Denza-GS yield fragmented or over-smoothed structures. All of them show blurred missing-wedge regions in their predicted sinograms. NanoMorph-3D recovers coherent 3D structures and completes the missing frequencies, matching the ground truth most closely.}
    \Description{Qualitative comparison of reconstructions and sinograms on four synthetic archetypes.}
    \label{fig:master_qualitative}
    \vspace{-4mm}
\end{figure*}

\begin{figure*}[t]
    \centering
    \includegraphics[width=\textwidth]{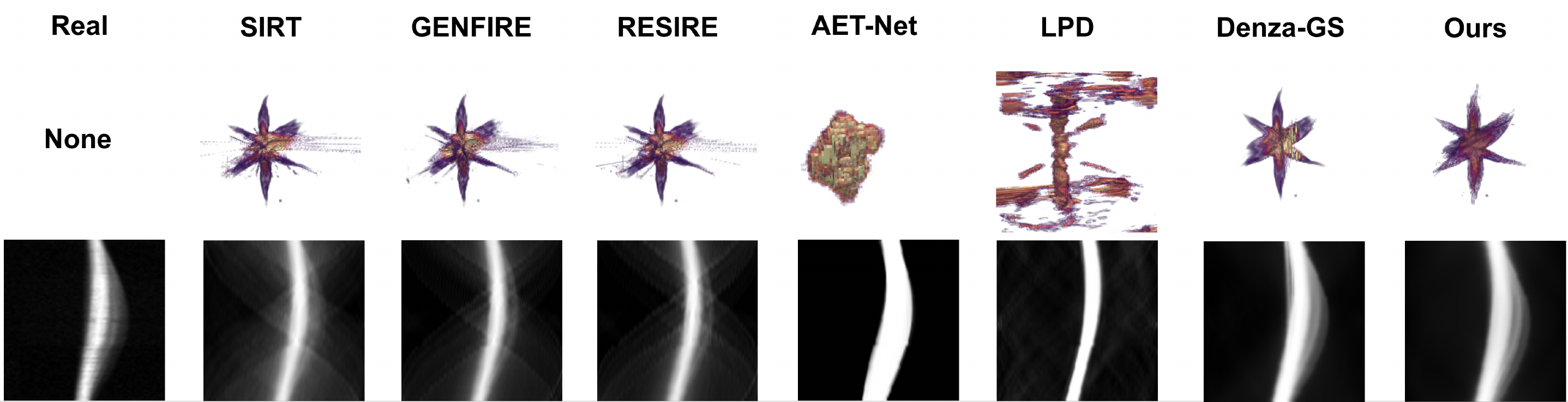}
    \vspace{-2mm}
    \caption{\textbf{Qualitative evaluation on real experimental data.} Reconstructed volumes are shown on top and their sinograms below. The baselines are degraded by experimental noise and yield blurred or structurally inconsistent volumes, whereas NanoMorph-3D recovers sharper, topologically consistent structures while suppressing instrumental noise.}
    \Description{Qualitative comparison of reconstructions and sinograms on real HAADF-STEM data.}
    \label{fig:real_qualitative}
    \vspace{-4mm}
\end{figure*}

Conversely, lacking ground-truth missing wedges in the unsupervised stream, the network leverages dynamically dropped views as pseudo-targets to enforce a Reprojection Consistency Loss ($\mathcal{L}_{\text{reproj}}$) against the original tilts. This curriculum-driven view synthesis suppresses Z-axis elongation artifacts, ensuring the reconstructed volume remains mathematically and physically faithful to the acquisition process across both domains.

\subsection{Synthesizing Nanomaterial Morphologies}
\label{subsec:dataset_simulation}
Existing learning-based reconstruction methods often suffer from generalization degradation due to morphological bias in their training data. To encompass the structural diversity of real-world nanomaterials, we introduce a Nanomorphological Taxonomy that moves beyond geometric primitives, abstracting physical syntheses into five primary archetypes: \textit{dense materials, layered structures, hollow materials, 1D structures, and porous materials}.

Guided by this taxonomy, we develop a large-scale physics-grounded synthetic dataset. We employ a GPU-accelerated procedural generation pipeline that decouples macroscopic topology from microscopic textures. For instance, amorphous structures are synthesized via smoothed continuous noise while crystalline lattices are modeled using Gyroid fields. To break the assumption of homogeneous interiors and prevent the network from learning binary mappings, we introduce continuous internal density fluctuations. We simulate heterogeneities, such as grain boundaries and crystal defects, by modulating the volumetric field with localized 3D Gaussian noise, constraining the density variations to a $\pm 20\%$ amplitude. This dual-level generation ensures our synthetic volumes accurately reflect the internal architecture of actual nanomaterials.

To adhere to the forward physical model defined in Sec.~\ref{subsec:problem_formulation}, we simulate the HAADF-STEM acquisition process over the generated volumes. Rather than relying on linear projections, we map the voxel grids to a physical scale of 100 nm and enforce a non-linear exponential attenuation governed by the Beer-Lambert law. We enhance robustness by applying domain randomization to the material attenuation coefficient $\mu$. Finally, we incorporate instrumental degradations such as electron-dose-dependent Poisson shot noise, background vacuum noise, and spatial detector blurring. By intertwining structural randomness with measurement degradations, this forward simulation provides a faithful testbed, ensuring that the morphology-agnostic priors learned by NanoMorph-3D are grounded in physical reality and capable of seamless simulation-to-reality transfer.

\begin{table*}[t]
  \caption{\textbf{Quantitative evaluation on the synthetic dataset.} Evaluation includes 3D volumes, observed views within $\pm 60^\circ$, and novel views in the missing wedge. NanoMorph-3D achieves higher spatial fidelity and consistency than compared methods across all observed and unmeasured projection regions.}
  \label{tab:synthetic_results}
  \centering
  \small
  \begin{tabular*}{\textwidth}{@{\extracolsep{\fill}} lccccccccc}
    \toprule
    \multirow{2}{*}{Method} & \multicolumn{3}{c}{3D Volume Metrics} & \multicolumn{3}{c}{2D Observed Views ($\pm 60^\circ$)} & \multicolumn{3}{c}{2D Novel Views (Missing Wedge)} \\
    \cmidrule(lr){2-4} \cmidrule(lr){5-7} \cmidrule(lr){8-10}
    & PSNR $\uparrow$ & SSIM $\uparrow$ & FSC $\uparrow$ & PSNR $\uparrow$ & SSIM $\uparrow$ & LPIPS $\downarrow$ & PSNR $\uparrow$ & SSIM $\uparrow$ & LPIPS $\downarrow$ \\
    \midrule
    
    SIRT~\cite{trampert1990simultaneous} & 22.46 & 0.4563 & 0.2430 & 38.15 & 0.8501 & 0.1031 & 32.89 & 0.7734 & 0.2418 \\
    GENFIRE~\cite{pryor2017genfire} & 23.05 & 0.5744 & 0.2569 & 40.85 & 0.9078 & 0.0898 & 33.42 & 0.8029 & 0.1844 \\
    RESIRE~\cite{pham2023accurate} & 24.49 & 0.4939 & 0.2701 & 42.27 & 0.9522 & 0.0827 & 34.60 & 0.8379 & 0.1706 \\
 
    AET-Net~\cite{lee2021single} & 15.29 & 0.3812 & 0.4529 & 22.10 & 0.6314 & 0.2747 & 18.23 & 0.5615 & 0.3469 \\ 

    LPD~\cite{adler2018learned} & 26.61 & 0.8934 & 0.3412 & 40.47 & 0.9053 & 0.1022 & 33.28 & 0.8190 & 0.1557 \\

    Denza-GS~\cite{zhang20263d} & 28.14 & 0.9125 & 0.4892 & 41.98 & 0.9282 & 0.0856 & 34.99 & 0.8591 & 0.1517 \\
    \midrule
    \rowcolor[gray]{.95}
    \textbf{NanoMorph-3D (Ours)} & \textbf{31.42} & \textbf{0.9518} & \textbf{0.6125} & \textbf{44.15} & \textbf{0.9705} & \textbf{0.0612} & \textbf{38.65} & \textbf{0.9240} & \textbf{0.0815} \\
    \bottomrule
  \end{tabular*}
\end{table*}

\begin{table}[t]
  \caption{\textbf{Reprojection consistency on experimental data.} Since 3D ground truth is unavailable for experimental tomograms, we evaluate performance using 2D reprojection PSNR, SSIM, and LPIPS across measured tilt angles to quantify structural and perceptual fidelity.}
  \label{tab:real_world_lpips}
  \centering
  \small 
  \setlength{\tabcolsep}{4pt} 
  \renewcommand{\arraystretch}{1.1} 
  \begin{tabular}{lccc}
    \toprule
    Method & PSNR $\uparrow$ & SSIM $\uparrow$ & LPIPS $\downarrow$ \\
    \midrule
    SIRT~\cite{trampert1990simultaneous}    & 32.15 & 0.764 & 0.284 \\
    GENFIRE~\cite{pryor2017genfire} & 33.42 & 0.812 & 0.215  \\
    RESIRE~\cite{pham2023accurate}  & 34.68 & 0.845 & 0.182  \\
    AET-Net~\cite{lee2021single} &  17.29   & 0.483 & 0.491 \\
    LPD~\cite{adler2018learned}     & 15.04 & 0.462 & 0.358 \\
    Denza-GS~\cite{zhang20263d} & 35.85 & 0.884 & 0.136  \\
    \midrule
    \rowcolor[gray]{.95}
    \textbf{NanoMorph-3D} & \textbf{39.24} & \textbf{0.941} & \textbf{0.078} \\
    \bottomrule
  \end{tabular}
\end{table}

\subsection{Synthetic-to-Reality Generalization}
\label{subsec:sim_to_real}
Although our taxonomy-driven synthetic dataset incorporates physical constraints, a domain gap remains between simulated data and experimental measurements. To bridge this gap without paired ground truth, we utilize a dual-stream consistency training strategy. NanoMorph-3D processes data through two parallel pathways with shared hierarchical volumetric operator weights. The supervised stream learns mesoscale geometry from synthetic sinograms and topological ground truth. Simultaneously, the unsupervised stream processes unlabeled experimental projections $Y_{\text{real}}$. 

We split the measured angles into two disjoint subsets $\Theta{\text{in}}$ and $\Theta{\text{out}}$, and reconstruct $V{\text{real}} = \text{Net}(Y{\Theta{\text{in}}})$ from the input subset alone. We then enforce a held-out reprojection loss on the withheld angles:

\begin{equation}
\mathcal{L}_{\text{unsup}} = \|\mathcal{A}_{\Theta_{\text{out}}}[V_{\text{real}}] - Y_{\Theta_{\text{out}}}\|_1
\end{equation}

Because $\Theta{\text{out}}$ never enters the reconstruction, minimizing this loss forces $V{\text{real}}$ to generalize to held-out angles rather than merely refit its input.

Sharing weights across both streams lets the network learn domain-invariant topological priors while adapting to experimental artifacts such as vacuum noise and residual scattering, thereby improving generalization to real data.

\section{Experiments}
\label{sec:experiments}

\subsection{Experimental Setup}
\label{subsec:setup}

\noindent\textbf{Datasets.} Due to the prohibitive cost of acquiring large-scale electron tomography data, we constructed a synthetic dataset of 6,000 3D volumes following the physics-based simulation pipeline described in Sec.~\ref{subsec:dataset_simulation}. For each volume, we generated full-tilt HAADF-STEM projections from $-90^\circ$ to $90^\circ$ at $3^\circ$ increments, incorporating instrumental degradations. The dataset is split into 5,000 for training, 500 for validation, and 500 for testing, with network inputs constrained to a limited-angle subset between $-60^\circ$ and $60^\circ$. To bridge the domain gap, we assembled seven real-world HAADF-STEM tilt series: three from a public nanomaterial database~\cite{levin2016nanomaterial} and four acquired in-house. The real series are used as unlabeled experimental data in the unsupervised domain-adaptation stream during training.

\noindent\textbf{Compared Methods.} We evaluate NanoMorph-3D against four reconstruction paradigms: the classical iterative solvers SIRT~\cite{trampert1990simultaneous}, GENFIRE~\cite{pryor2017genfire}, and RESIRE~\cite{pham2023accurate}, the image-domain post-processing method AET-Net~\cite{lee2021single}, the physics-driven unrolled network LPD~\cite{adler2018learned}, and the 3D neural rendering approach Denza-GS~\cite{zhang20263d}.

\noindent\textbf{Evaluation Metrics.} Evaluation is conducted across our synthetic dataset and real-world experimental data. On synthetic dataset, 3D ground truth enables a dual-domain assessment as reported in Table~\ref{tab:synthetic_results}. We measure spatial fidelity and high-frequency recovery via PSNR~\cite{jahne2005digital}, SSIM~\cite{wang2004image}, and Fourier Shell Correlation (FSC)~\cite{van2005fourier}. To prevent bias toward over-smoothed results, 2D reprojection metrics, notably LPIPS~\cite{zhang2018unreasonable}, are included to assess structural fidelity. For uncalibrated real-world HAADF-STEM tilt series, where 3D ground truth is unavailable, performance is assessed via visual topological fidelity and 2D reprojection consistency using PSNR~\cite{jahne2005digital}, SSIM~\cite{wang2004image}, and LPIPS~\cite{zhang2018unreasonable} in Table~\ref{tab:real_world_lpips}.

\noindent\textbf{Implementation Details.} Our framework is implemented in PyTorch and the PGD algorithm is unrolled for $K=10$ stages, with the loss balancing weight $\lambda$ set to $0.1$. To mitigate memory constraints during high-resolution 3D Transformer computations, we employ gradient checkpointing and gradient accumulation with an effective batch size of $1$. The network is optimized using AdamW with an initial learning rate of $2 \times 10^{-4}$ and a cosine annealing schedule. All experiments and training processes are conducted on four NVIDIA RTX 3090 GPUs.

\subsection{Comparison Experiments}
\label{subsec:sota}

\noindent\textbf{Quantitative Comparison.} Table~\ref{tab:synthetic_results} summarizes the evaluation on the synthetic dataset, where dense 3D ground truth is available. Traditional iterative solvers (SIRT, GENFIRE, and RESIRE) struggle to reconstruct complex structures, as evidenced by their low FSC scores (all below 0.30), which indicate severe over-smoothing of high-frequency details. The image-domain network AET-Net suffers a drastic drop in spatial fidelity (yielding a PSNR of 15.29), confirming its inability to preserve volumetric consistency without explicit physical constraints. While the unrolled network LPD achieves a higher SSIM, its reliance on local 2D receptive fields hinders accurate 3D topology recovery within the missing wedge. Although Denza-GS serves as a competitive neural rendering method, NanoMorph-3D outperforms all methods by a large margin, achieving the highest FSC (0.6125) and lowest LPIPS, effectively resolving missing wedge hallucinations. 
Furthermore, Table~\ref{tab:real_world_lpips} evaluates generalization on experimental HAADF-STEM tilt series via 2D reprojection consistency. Under uncalibrated real-world conditions with instrumental noise, data-driven methods like AET-Net and LPD experience catastrophic degradation, with reprojection PSNRs falling below 18 dB. In contrast, NanoMorph-3D maintains robust performance, yielding the highest reprojection PSNR (39.24) and lowest LPIPS (0.078). By neither amplifying background artifacts nor losing fine structural details, our framework demonstrates successful simulation-to-reality transfer, providing reliable, high-fidelity 3D characterization for unlabeled experimental data.

\noindent\textbf{Qualitative Comparison.} As illustrated in Fig.~\ref{fig:master_qualitative} and Fig.~\ref{fig:real_qualitative}, visual comparisons corroborate our quantitative findings. Under limited-angle conditions, traditional iterative methods (SIRT, GENFIRE, and RESIRE) suffer from severe Z-axis elongation and blurring artifacts caused by the missing wedge, frequently collapsing the delicate internal voids of nanomaterials. The image-domain method AET-Net struggles to maintain structural integrity, yielding fragmented topologies on the synthetic dataset and completely failing to recover the underlying morphology on real inputs, as its refinement process is decoupled from explicit physical constraints. While the unrolled LPD maintains reasonable visual coherence on synthetic data, its reliance on local 2D receptive fields prevents it from generalizing to uncalibrated real-world conditions. On experimental data, LPD exhibits severe slice-wise repeating artifacts and incorrect 3D voxel distributions. Furthermore, neural rendering methods like Denza-GS generate noisy boundaries and fail to correctly resolve continuous internal cavities. In contrast, NanoMorph-3D accurately recovers the full morphological spectrum. Notably, in complex porous frameworks, our method successfully preserves pore-channel connectivity, eliminates Z-axis stretching, and yields sharp, physically consistent boundaries. This confirms that explicitly embedding physical projection operators alongside our Dual-Domain Sinusoidal Attention (DDSA) effectively resolves the missing wedge ambiguity across both simulated and experimental domains.

\begin{table}[htbp]
  \caption{Quantitative comparison of 3D reconstruction metrics on the synthetic dataset under sparse projection views.}
  \label{tab:sparse_metrics}
  \centering
  \resizebox{\linewidth}{!}{
  \begin{tabular}{l|cc|cc|cc|cc}
    \toprule
    \multirow{2}{*}{\textbf{Method}} & \multicolumn{2}{c|}{\textbf{41 Views}} & \multicolumn{2}{c|}{\textbf{31 Views}} & \multicolumn{2}{c|}{\textbf{21 Views}} & \multicolumn{2}{c}{\textbf{11 Views}} \\
    & SSIM $\uparrow$ & FSC $\uparrow$ & SSIM $\uparrow$ & FSC $\uparrow$ & SSIM $\uparrow$ & FSC $\uparrow$ & SSIM $\uparrow$ & FSC $\uparrow$ \\
    \midrule
    SIRT~\cite{trampert1990simultaneous} & 0.456 & 0.243 & 0.382 & 0.212 & 0.291 & 0.175 & 0.185 & 0.124 \\
    GENFIRE~\cite{pryor2017genfire} & 0.574 & 0.257 & 0.495 & 0.231 & 0.388 & 0.198 & 0.254 & 0.145 \\
    RESIRE~\cite{pham2023accurate} & 0.494 & 0.270 & 0.425 & 0.245 & 0.332 & 0.208 & 0.215 & 0.152 \\
    AET-Net~\cite{lee2021single} & 0.381 & 0.453 & 0.314 & 0.375 & 0.228 & 0.280 & 0.155 & 0.185 \\
    LPD~\cite{adler2018learned} & 0.893 & 0.341 & 0.835 & 0.295 & 0.720 & 0.240 & 0.510 & 0.160 \\
    Denza-GS~\cite{zhang20263d} & 0.912 & 0.489 & 0.865 & 0.442 & 0.758 & 0.368 & 0.540 & 0.245 \\
    \midrule
    \rowcolor[gray]{.95}
    \textbf{NanoMorph-3D} & \textbf{0.951} & \textbf{0.612} & \textbf{0.938} & \textbf{0.585} & \textbf{0.915} & \textbf{0.542} & \textbf{0.888} & \textbf{0.495} \\
    \bottomrule
  \end{tabular}
  }
\end{table}

\subsection{Robustness to Sparse Views}Dense tilt series acquisition causes beam damage to nanomaterials, making view reduction essential for low-dose imaging. However, sparse sampling exacerbates the informational deficiency for 3D recovery. We evaluate NanoMorph-3D on the entire synthetic dataset by subsampling the $\pm 60^{\circ}$ tilt range into 41, 31, 21, and 11 views.

As shown in Table~\ref{tab:sparse_metrics}, traditional solvers and image-domain networks experience severe performance degradation as views decrease. At the extreme sparsity of 11 views, SIRT and GENFIRE SSIM values plummet to 0.185 and 0.254 respectively, while Denza-GS falls to 0.540. In contrast, NanoMorph-3D maintains a high SSIM of 0.888 and an FSC of 0.495. Notably, NanoMorph-3D with only 11 views achieves higher fidelity than GENFIRE using the full 41-view sequence. Our framework effectively navigates the expanded null space, outperforming GENFIRE operating on the full 41 views and highlighting the resilience of embedding physical forward models into deep architectures.

\begin{table}[t] 
  \centering
  \caption{\textbf{Ablation Study on the Synthetic Dataset.} We validate our proposed modules by systematically removing or replacing them from the full NanoMorph-3D network and evaluating on the entire test set. Removing the explicitly unrolled Physical Gradient Scaling (PGS) causes the most significant structural degradation, while removing the Dual-Domain Sinusoidal Attention (DDSA) leads to a loss of high-frequency details indicated by FSC.}
  \label{tab:ablation}
  \resizebox{\columnwidth}{!}{
  \begin{tabular}{ccc|ccc}
    \toprule
    \textbf{PGS} & \textbf{PhysNorm} & \textbf{DDSA} & \textbf{PSNR $\uparrow$} & \textbf{SSIM $\uparrow$} & \textbf{FSC $\uparrow$} \\
    \midrule
    \rowcolor[gray]{.95}
    \checkmark & \checkmark & \checkmark & \textbf{31.42} & \textbf{0.9518} & \textbf{0.6125} \\
    \checkmark & \checkmark & \texttimes & 30.15 & 0.9320 & 0.5410 \\
    \checkmark & \texttimes & \texttimes & 28.90 & 0.9105 & 0.4820 \\
    \texttimes & \texttimes & \texttimes & 26.45 & 0.8712 & 0.3540 \\
    \bottomrule
  \end{tabular}
  }
\end{table}

\subsection{Ablation Study}
\label{subsec:ablation}
We conduct an ablation study to systematically validate the contribution of each core innovation in NanoMorph-3D. Table~\ref{tab:ablation} reports the quantitative degradation on the synthetic dataset as specific components are systematically removed from the full model.

\noindent\textbf{Effectiveness of Dual-Domain Sinusoidal Attention.} Removing the Dual-Domain Sinusoidal Attention (DDSA) deprives the network of its geometric inductive bias, so it can no longer propagate features along projection trajectories into undersampled regions. FSC drops from 0.6125 to 0.5410 and PSNR from 31.42 to 30.15.

\noindent\textbf{Impact of Physics-Normalization.} Substituting standard Layer Normalization treats absolute physical scales as statistical variance, so the network loses scale-invariant feature extraction across differing magnifications and mass densities. PSNR falls to 28.90 and SSIM to 0.9105.

\noindent\textbf{Effectiveness of PGD Unrolling.} Completely reverting to a black-box 3D network removes the explicitly unrolled Physical Gradient Scaling (PGS) within the Projection Consistency Unit. Decoupling the reconstruction process from physical measurements eliminates intermediate data consistency constraints. The model consequently suffers the most substantial structural degradation, with PSNR plummeting to 26.45 and FSC falling to 0.3540. This confirms that physics-grounded reconstruction of 3D structures is essential for suppressing topological artifacts.

\section{Conclusion}
\label{sec:conclusion}
NanoMorph-3D is an unrolling framework for electron tomography reconstruction that addresses the missing wedge problem in nanomaterials. By integrating the Proximal Gradient Descent algorithm with Dual-Domain Sinusoidal Attention and Physics-Normalization, the method combines physical projection models with morphological priors. Evaluations demonstrate that the framework preserves pore connectivity and suppresses Z-axis elongation. It outperforms compared methods in structural fidelity and inference speed. A dual-stream consistency strategy enables domain transfer from simulation to reality on experimental data. Although constrained by the memory footprint of 3D Transformers and reliance on the Beer-Lambert law, NanoMorph-3D provides a physics-driven approach for the 3D characterization of nanomaterials.

\begin{acks}
\sloppy
This work was supported by the National Natural Science Foundation of China (Grant Nos. 62331006 and 12274025) and the Beijing Science and Technology Planning Project (Grant No. Z251100006925032).

\end{acks}

\bibliographystyle{ACM-Reference-Format}
\bibliography{sample-base}

\end{document}